\documentclass[11pt]{article}
\usepackage{acl2013}
\usepackage{times}
\usepackage{url}
\usepackage{latexsym}
\usepackage{graphicx}
\usepackage{epstopdf}
\usepackage{subfigure}
\usepackage{multirow}
\usepackage{paralist}
\usepackage{CJKutf8}
\usepackage{amsmath}
\usepackage{epstopdf}
\usepackage{tikz}
\usepackage{amssymb}
\usepackage{amsmath}
\usepackage{algorithmic}
\usepackage{algorithm}
\usepackage{subfigure}
\usepackage{multirow}
\usepackage{bm}
\usepackage{array}
\usepackage{placeins}
\usepackage{multirow}

\title{Radical-Enhanced Chinese Character Embedding}

\author{Yaming Sun$^{\dag}$, Lei Lin$^{\dag}$ , Duyu Tang$^{\dag}$, Nan Yang$^{\ddag}$, Zhenzhou Ji$^{\dag}$, Xiaolong Wang$^{\dag}$ \\
$^{\dag}$Harbin Institute of Technology, China \\
$^{\ddag}$University of Science and Technology of China, Hefei, China
}

\date{}

\begin{document}

\maketitle
\begin{abstract}
We present a method to leverage radical for learning Chinese character embedding. \textbf{Radical} is a semantic and phonetic component of Chinese character. It plays an important role as characters with the same radical usually have similar semantic meaning and grammatical usage. However, existing Chinese processing algorithms typically regard word or character as the basic unit but ignore the crucial radical information. In this paper, we fill this gap by leveraging radical for learning continuous representation of Chinese character. We develop a dedicated neural architecture to effectively learn character embedding and apply it on Chinese character similarity judgement and Chinese word segmentation. Experiment results show that our radical-enhanced method outperforms existing embedding learning algorithms on both tasks.
\end{abstract}
\begin{CJK*}{UTF8}{gbsn}
\section{Introduction}
Chinese ``\textbf{radical} (部首)" is a graphical component of Chinese character, which serves as an indexing component in the Chinese dictionary\footnote{http://en.wikipedia.org/wiki/Radical\_(Chinese\_character)}. In general, a Chinese character is phono-semantic, with a radical as its semantic and phonetic component suggesting part of its meaning. For example, ``氵(water)" is the radical of ``河~(river)", and ``足~(foot)" is the radical of ``跑~(run)".

Radical is important for the computational processing of Chinese language. The reason lies in that characters with the same radical typically have similar semantic meanings and play similar grammatical roles. For example, verbs ``打 (hit)" and ``拍 (pat)" share the same radical ``扌(hand)" and usually act as the subject-verb in sentences. To our best knowledge, existing studies in Chinese NLP tasks, such as word segmentation, typically treat word \cite{Zhang2010} or character \cite{Zhang2013} as the basic unit, while ignore the radical information. In this paper, we leverage the radical information of character for the computational processing of Chinese. Specifically, we exploit the radical of character for learning Chinese character embedding.
\end{CJK*}
Most existing embedding learning algorithms~\cite{Bengio2003,Morin2005,Mikolov2010,Huang2012,Luong2013,Mikolov2013} model the representation for a word with its context information.
We extend an existing embedding learning algorithm~\cite{Collobert2008,Collobert2011} and propose a tailored neural architecture to leverage radical for learning the continuous representation of Chinese character. Our neural model integrates the radical information by predicting the radical of each character through a $softmax$ layer. Our loss function is the linear combination of the loss of C\&W model \cite{Collobert2011} and the cross-entropy error of $softmax$. 
We apply the radical-enhanced character embedding on two tasks, Chinese character similarity judgement and Chinese word segmentation. 
Experiment results on both tasks show that, our method outperforms existing embedding learning algorithms which do not utilize the radical information.
\begin{figure*}[!t]
\begin{center}
\includegraphics[scale=1.7,width=1.7\columnwidth]{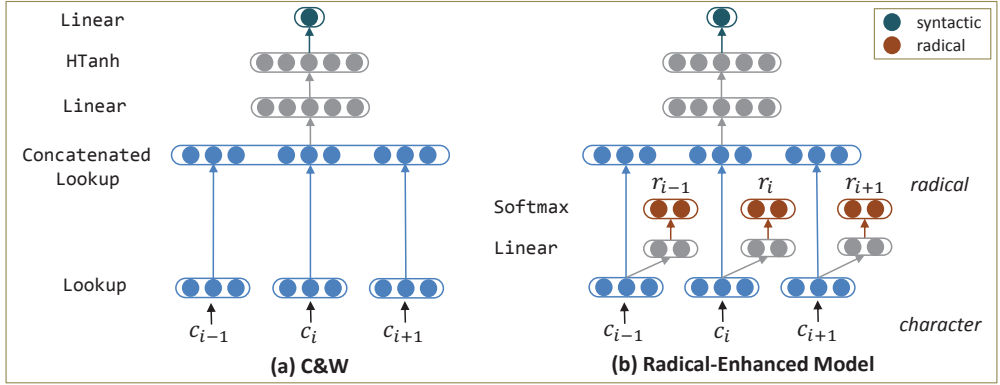}
\caption{Architecture of C\&W (a) and our radical-enhanced character embedding learning model (b).}
\label{Figure_architecture} 
\end{center}
\end{figure*}
The major contributions of this paper are summarized as follows.
\begin{itemize}
\item To our best knowledge, this is the first work that leverages radical for learning Chinese character embedding.
\item We learn Chinese character embedding by exploiting the radical information of character and verify its effectiveness on two tasks.
\item We publish the radical-enhanced Chinese character embedding, which can be easily applied on other Chinese NLP tasks. We also introduce a dataset on Chinese character similarity judgement.
\end{itemize}

This paper is organized as follows. We propose the radical-enhanced character representation learning model in Section \ref{approach}. In section 4, we introduce the Chinese word segmentation task and the neural Conditional Random Field (CRF) model for utilizing character embedding as features. Then we present the experimental studies in Section 5, and finally conclude the paper in Section 6.

\section{Related Work}
\label{related work}
In this section, we review the previous studies from two directions, namely learning word embedding and applying word embedding for NLP applications.

\subsection{Learning Word Embedding}
It is well-accepted that the representation of word is the basis of the field of natural language processing \cite{Turney2010,Turian2010}. In the early studies, a word is represented as a one-hot vector, whose length is the size of vocabulary, and only one dimension is 1, others are 0. The main drawback of the one-hot representation is that it can not reflect the grammatical and semantic relations between words. To overcome this shortcoming, some studies have been done to learn the latent factors of words, such as Latent Semantic Analysis \cite{Deerwester1990} and Latent Dirichlet Allocation \cite{Blei2003}.
With the revival of deep learning \cite{Bengio2013}, many researchers focus on the continuous representation of words (a.k.a word embedding).
Existing embedding learning algorithms can be divided into two directions based on the use of unstructured raw texts~\cite{Collobert2011} or structured knowledge base~\cite{Bordes2011}. Due to the lack of large-scale Chinese knowledge base (KB), this paper focuses on learning character embedding from unstructured corpus and leaves the KB-based method to the future work.
From the perspective of learning embedding from raw corpus, most existing algorithms model the representation for a word with its context information. \newcite{Bengio2003} propose a feed-forward neural probabilistic language model to predict the next word based on its previous contextual words. Based on their work, some methods are presented to reduce the training time of neural language model. \newcite{Morin2005} and \newcite{mnih2008scalable} propose hierarchical language models, which encode the vocabulary-sized $softmax$ layer into a tree structure. \newcite{Collobert2008} propose a feed-forward neural network (C\&W) which learns word embedding with a ranking-type cost. Mikolov et al. introduce the Recurrent neural network language models (RNNLMs) \cite{Mikolov2010}, Continuous Bag-of-Word (CBOW) and skip-gram model \cite{Mikolov2013b} to learn embedding for words and phrases. \newcite{Huang2012} propose a neural model to utilize the global context in addition to the local information. Besides utilizing neural networks to learn word embedding, some recent studies try the PCA-based algorithm to simplify the computation process \cite{Lebret2013}.
The representation of words heavily relies on the characteristic of language. The linguistic feature of English has been studied and used in the word embedding learning procedure. Specifically, \newcite{Luong2013} utilize the morphological property of English word and incorporate the morphology into word embedding. In this paper, we focus on learning Chinese character embedding by exploiting the radical information of Chinese character, which is tailored for Chinese language. Unlike \newcite{Luong2013} that initialize their model with the pre-trained embedding, we learn Chinese character embedding from scratch.
\subsection{Word Embedding for NLP Tasks}

Word embedding is able to capture the syntactic and semantic meanings of a word from massive corpora, which can reflect the discriminative features of data. Recently, word embedding has been successfully applied to a variety of NLP tasks, such as chunking, named entity recognition \cite{Turian2010}, POS tagging, semantic role labeling \cite{Collobert2011}, sentiment analysis \cite{Socher2013a}, paraphrase detection \cite{Socher2011a}, parsing \cite{Socher2013} and Chinese word segmentation \cite{Mansur2013,Zheng2013}. For the task of Chinese word segmentation, \newcite{Mansur2013} propose a feature-based neural language model for learning feature embedding. They develop a deep neural architecture which takes the embedding as input and tag the sequence. \newcite{Zheng2013} present a neural architecture which combines embedding learning and sequence tagging in a unified model. The two studies on Chinese word segmentation utilize character embedding, yet they do not take the radical nature of Chinese language into consideration. Unlike previous studies, we leverage the radical information into the embedding learning process.

In this paper, we propose a neural network architecture tailored for Chinese character representation learning utilizing the radical information which is an typical characteristic of Chinese. We apply the learned embedding into a neural-CRF based Chinese word segmentation framework \cite{Zheng2013} to verify its effectiveness. Neural-CRF is a sequential labeling framework that incorporates the representations of word (or character) into the CRF with a feed-forward neural network (detailed in Section \ref{CRF}).
In the neural-CRF model, the word (or character) embeddings are treated as input features and the performance of further application highly depends on the quality of word (or character) representation.

\section{Radical-Enhanced Model for Chinese Character Representation Learning}
\label{approach}
In this section, we describe the details of leveraging the radical information for learning Chinese character embedding. Based on C\&W model~\cite{Collobert2011}, we present a radical-enhanced model, which utilizes both radical and context information of characters. In the following subsections, we first briefly introduce the C\&W model, and then present the details of our radical-enhanced neural architecture.

\subsection{C\&W Model}
C\&W model ~\cite{Collobert2011} is proposed to learn the continuous representation of a word from its context words. Its training objective is to assign a higher score to the reasonable ngram than the corrupted ngram. The loss function of C\&W is a ranking-type cost:
\begin{equation}
\label{loss-cw}
loss_{c}(s,s^{w})= max(0,1-score(s)+score(s^{w}))
\end{equation}
where $s$ is the reasonable ngram, $s^{w}$ is the corrupted one with the middle word replaced by word $w$, and $score(.)$ represents the reasonability scalar of a given ngram, which can be calculated by its neural model.

C\&W is a feed-forward neural network consisted of four layers, as illustrated in Figure \ref{Figure_architecture}(a). The input of C\&W is a ngram composed of $n$ words, and the output is a score which evaluates the reasonability of the ngram. Each word is encoded as a column vector in the embedding matrix $\mathbf{W}_{e}\in \mathbb{R}^{d\times \mid V\mid}$, where $d$ is the dimension of the vector, and $V$ is the vocabulary. The $lookup$ layer has a fixed window size $n$, and it maps each word of the input ngram into its embedding representation. The output $score(s)$ is computed as follows:
\begin{equation}
score(s) = \mathbf{W}_{2}\mathbf{a}+b_{2}
\end{equation}
\begin{equation}
\mathbf{a} = HTanh(\mathbf{W}_{1}[\mathbf{x}_{1}...\mathbf{x}_{n}]+\mathbf{b}_{1})
\end{equation}
where $[\mathbf{x}_{1}...\mathbf{x}_{n}]$ is the concatenation of the embedding vectors of words $x_{1},...x_{n}$, $\mathbf{W}_{1},\mathbf{W}_{2},\mathbf{b}_{1},b_{2}$ are the weights and biases of the two linear layers, and function $HTanh(.)$ is the $HardTanh$ function.
The parameters can be learned by minimizing the loss through stochastic gradient descent algorithm.
\begin{figure*}[t]
\begin{center}
\includegraphics[scale=1.8,width=1.8\columnwidth]{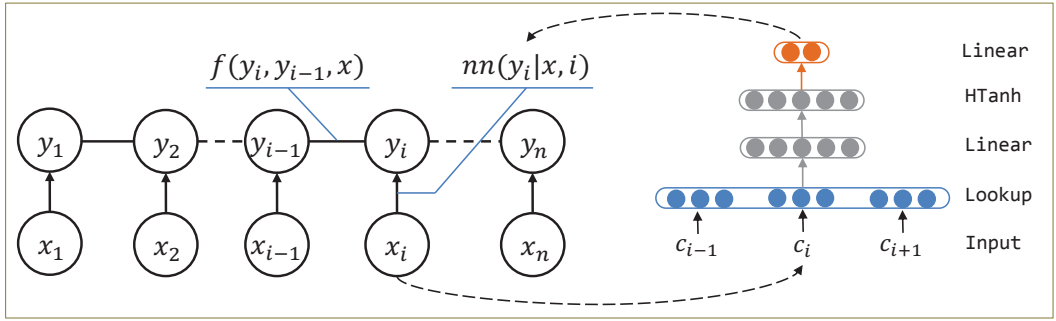}
\vspace{-1em}
\caption{The neural CRF for Chinese word segmentation. Each input character $x_{i}$ is denoted with its embedding vector, and $window(x_{i})$ is the input of the neural network.}
\label{Figure_crf} 
\end{center}
\end{figure*}
\subsection{Radical-Enhanced Model}

In this part, we present the radical-enhanced model for learning Chinese character embedding. Our model captures the radical information as well as the context information of characters. 

The training objective of our radical-enhanced model contains two parts: 1) for a ngram, discriminate the correct middle character from the randomly replaced character; 2) for each character within a ngram, predict its radical. To this end, we develop a tailored neural architecture composed of two parts, context-based part and radical-based part, as given in Figure \ref{Figure_architecture}(b).
The context-based part captures the context information, and the radical-based part utilizes the radical information. The final loss function of our model is shown as follows:
\begin{equation}
\label{equation-combine}
\begin{aligned}
L&oss(s, s^w)\ = \alpha \cdot  loss_{c}(s,s^{w}) + \\&(1-\alpha) \cdot (\sum_{c\in s}loss_{r}(c)+\sum_{c\in s^{w}}loss_{r}(c))
\end{aligned}
\end{equation}
where $s$ is the correct ngram, $s^{w}$ is the corrupted ngram, $loss_{c}(.)$ is the loss of the context-based part, $loss_{r}(.)$ is the loss of the radical-based part, and $\alpha$ linearly weights the two parts.

Specifically, the context-based part takes a ngram as input and outputs a score, as described in Equation \ref{loss-cw}.
The radical-based part is a list of feed-forward neural networks with shared parameters, each of which is composed of three layers, namely $lookup \rightarrow linear \rightarrow softmax$ (from bottom to top). The unit number of each $softmax$ layer is equal to the number of radicals. Softmax layer is suitable for this scenario as its output can be interpreted as conditional probabilities. The cross-entropy loss of each softmax layer is defined as follows:
\begin{equation}
\label{loss-radical}
loss_r(c) = -\sum_{i = 0}^{N}\bm{p}_{i}^{g}(c) \times log(\bm{p}_{i}(c))\
\end{equation}
where $N$ is the number of radicals; $\bm{p}^g(c)$ is the gold radical distribution of character $c$, with $\sum_{i} \bm{p}_i^g(c) = 1$; $\bm{p}_{}(c)$ is the predicted radical distribution.

\paragraph{Model Training}
Our model is trained by minimizing the loss given in Equation \ref{equation-combine} over the training set. The parameters are embedding matrix of Chinese characters, weights and biases of each linear layer. All the parameters are initialized with random values, and updated via stochastic gradient descent. Hyper-parameter $\alpha$ is tuned on the development set.

\section{Neural CRF for Chinese Word Segmentation}
\label{CRF}

It is widely accepted that Chinese Word Segmentation can be resolved as a character based tagging problem \cite{xue2003chinese}. In this paper, we treat word segmentation as a sequence tagging task, and assign characters with four possible boundary tags: ``B" for a character at the beginning of a word, ``I" for the characters inside a word, ``E" for that at the end of a word, and ``S" for the character which is a word itself \cite{zhengdeep}.
\subsection{Traditional CRF}
Linear chain conditional random field (CRF) \cite{lafferty2001conditional} is a widely used algorithm for Chinese word segmentation. Given an observation sequence $\vec{\mathbf{x}}$ and its gold tag sequence $\vec{\mathbf{y}}$, CRF models a conditional probability distribution as follows,
\begin{equation}
\label{conditional-prob1}
P(\vec{\mathbf{y}}|\vec{\mathbf{x}})=\frac{1}{Z}\prod_{C}\Psi_{c}(Y_{c})=\frac{exp\ \phi(\vec{\mathbf{y}}, \vec{\mathbf{x}})}
{\sum_{\vec{\mathbf{y}}'}^{}exp\ \phi(\vec{\mathbf{y}}', \vec{\mathbf{x}})}
\end{equation}
where $C$ is a maximum clique, $\Psi_{C}(Y_{C})$ is the potential function which is defined as an exponential function, $exp\phi(\vec{\mathbf{y}},\vec{\mathbf{x}})$ is the product of potential function on all the maximum cliques, and $Z$ is the normalization factor. Function $\phi(\vec{\mathbf{y}}, \vec{\mathbf{x}})$ is defined as follows:
\begin{equation}
\phi(\vec{\mathbf{y}}, \vec{\mathbf{x}})=\sum_{i,k}\lambda_{k}t_{k}(y_{i-1},y_{i},\vec{\mathbf{x}},i)+\sum_{i,l}\mu_{l}s_{l}(y_{i},\vec{\mathbf{x}},i)
\end{equation}
where $t_{k}$ and $s_{l}$ are feature functions, $\lambda_{k}$ and $\mu_{l}$ are the corresponding weights.

\subsection{Neural CRF}
In this section, we apply the radical-enhanced character embedding for Chinese word segmentation. Instead of hand-crafting feature, we leverage the learned character embedding as features for Chinese word segmentation with Neural CRF~\cite{Turian2010,Zheng2013}. The illustration of neural CRF is shown in Figure \ref{Figure_crf}. Given an observation sequence $\vec{\mathbf{x}}$ and its gold tag sequence $\vec{\mathbf{y}}$, neural CRF models their conditional probability as follows,

\begin{equation}
\label{conditional-prob}
P(\vec{\mathbf{y}}|\vec{\mathbf{x}})=\frac{exp\ \phi(\vec{\mathbf{y}}, \vec{\mathbf{x}})}
{\sum_{\vec{\mathbf{y}}'}^{}exp\ \phi(\vec{\mathbf{y}}', \vec{\mathbf{x}})}
\end{equation}
where $\phi(\vec{\mathbf{y}},\vec{\mathbf{x}})$ is the potential function which is computed as follows,
\begin{equation}
\label{potential}
\phi(\vec{\mathbf{y}}, \vec{\mathbf{x}})=\sum_{i}^{}
[f(y_i, y_{i-1}, \vec{\mathbf{x}})\vec{\mathbf{w}_1} + f(y_i, \vec{\mathbf{x}})\vec{\mathbf{w}_2}]
\end{equation}
where $f(y_i,y_{i-1},\vec{\mathbf{x}})$ is a binary-valued indicator function reflecting the transitions between $y_{i-1}$ and $y_i$, and $\vec{\mathbf{w}_1}$ is its associated weight.
$f(y_i,\vec{\mathbf{x}})\vec{\mathbf{w}_2}$ reflects the correlation of the input $\vec{\mathbf{x}}$ and the i-th label $y_i$, which is calculated by a four-layer neural network as given in Figure \ref{Figure_crf}. The neural network takes a ngram as input, and outputs a distribution over all possible tags, such as ``B/I/E/S". The unit number of the top $linear$ layer is equal to the number of tags, and the output is computed as follows,
\begin{equation}
\mathbf{output} = \mathbf{W}_{2}\mathbf{a}+\mathbf{b}_{2}
\end{equation}
\begin{equation}
\mathbf{a} = HTanh(\mathbf{W}_{1}\mathbf{window}(c_i)+\mathbf{b}_{1})
\end{equation}
\begin{equation}
\mathbf{window}(c_i) = [\mathbf{c}_{i-m}...\mathbf{c}_{i}...\mathbf{c}_{i+m}]
\end{equation}
where $c_i$ is the current character, $m$ is the window size, $\mathbf{window}(c_i)$ is the concatenation of the embeddings of $c_i$'s context characters, $\mathbf{W}_1,\mathbf{W}_2,\mathbf{b}_1,\mathbf{b}_2$ are the weights and biases of the linear layers, $HTanh$ is the $HardTanh$ function.

The neural CRF is trained via maximizing the likelihood of $P(\vec{\mathbf{y}}|\vec{\mathbf{x}})$ over all the sentences in the training set. We use Viterbi algorithm \cite{ForneyJr1973} in the decoding procedure.

\section{Experiments}
In this section, we evaluate the radical-enhanced character embedding on two tasks, Chinese character similarity judgement and Chinese word segmentation. We compare our model with C\&W \cite{Collobert2011} and word2vec\footnote{Available at https://code.google.com/p/word2vec/. We utilize Skip-Gram as baseline.}~\cite{Mikolov2013b}, and learn Chinese character embedding with the same settings.
\begin{CJK*}{UTF8}{gbsn}
To effectively train character embeddings, we randomly select one million sentences from the Sougou corpus\footnote{http://www.sogou.com/labs/dl/c.html}. We extract a radical mapping dictionary from an online Chinese dictionary\footnote{http://xh.5156edu.com/}, which contains 265 radicals and 20,552 Chinese characters. Each character listed in the radical dictionary is attached with its radical, such as $\langle$吃(eat), 口(mouth)$\rangle$.
\end{CJK*}
We empirically set the embedding size as 30, window size as 5, learning rate as 0.1, and the length of hidden layer as 30.

\subsection{Chinese Character Similarity}
  In this part, we evaluate the effectiveness of character embedding through Chinese character similarity judgement in the embedding space. Due to the lack of public dataset in Chinese, we build an evaluation dataset manually.
  
  In view of polysemy, we divide characters into different clusters according to their most frequently-used meanings. The dataset totally contains 26 categories and 988 characters. The evaluation metric is the accuracy of semantic consistency between each character and its top K nearest neighbors. The accuracy is calculated as follows,
\begin{equation}
\label{equation-accuracy}
Accuracy = \frac{1}{|S|}\sum_{c_{i} \in S}\frac{1}{K}\sum_{t_{j} \in top(c_{i})}\delta(c_{i},t_{j})
\end{equation}
where $S$ is the dataset, $c_{i}$ is a character, $top(c_{i})$ is the top $K$ nearest neighbors of $c_{i}$ in the embedding space using cosine similarity. $\delta(c_{i},t_{j})$ is an indicator function which is equal to 1 if $c_i$ and $t_j$ have the same semantic category, and equal to 0 on the contrary. We set $K$=$10$ in the following experiment.

\begin{figure}[htbp]
\begin{center}
\includegraphics[scale=0.95,width=0.95\columnwidth]{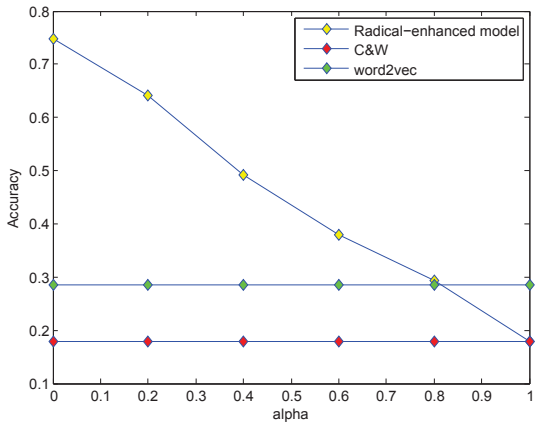}
\vspace{-1em}
\caption{Accuracy of C\&W, word2vec and our radical-enhanced model with K=10.}
\label{Figure_alpha} %
\end{center}
\end{figure}

\paragraph{Results and Analysis}
Figure \ref{Figure_alpha} shows the accuracy of our radical-enhanced model and baseline embedding learning algorithms on character similarity judgement.
The $alpha$ on the x-axis is the weight of the context-based component in our radical-enhanced model. Our model with $alpha$=1.0 represents the C\&W model.
Results show that our radical-enhanced model outperforms C\&W and word2vec consistently when $alpha$ is lower than 0.8. The reason is that our model can effectively leverage rich semantic information from radicals, which are not explicitly captured in the baseline embedding learning algorithms.
We also find that the accuracy of our model decreases with the increase of $alpha$ because the impact of radical is larger with smaller $alpha$. The trend further verifies the effectiveness of radical information.
\subsection{Chinese Word Segmentation}

In this part, we apply character embedding as features for Chinese word segmentation using neural CRF. We conduct experiments on the widely-used Penn Chinese Treebanks 5 (CTB5) and CTB7. CTB5 is split according to \cite{Jiang2008}. CTB7 is split according to~\cite{Wang2011}. 
The details of the datasets are given in Table \ref{Table-CTB}.

\begin{table}[!h]
\centering
	\begin{tabular}{|c|c|c|c|}
	\hline
	&Training&Development&Test\\ \hline
	CTB5&18,085&350&348 \\ \hline
    CTB7&31,088&10,036&10,291 \\
	\hline
	\end{tabular}
\caption{Statistics of the datasets of CTB5 and CTB7 for Chinese word segmentation.}\label{Table-CTB}
\end{table}

The parameters of the neural CRF are empirically set as follows, the window size is 3, the hidden layer is set with 300 units, and the learning rate is set to 0.1. The evaluation criterion is Precision ($P$), Recall ($R$) and F1-score ($F_1$).
\begin{table*}[t]
\centering
	\begin{tabular}{|l|ccc|ccc|}
	\hline
	\multirow{2}{*}{Method}&\multicolumn{3}{c|}{CTB5}&\multicolumn{3}{c|}{CTB7}\\
	\cline{2-7}
    &$P$&$R$&$F_1$&$P$&$R$&$F_1$\\
    \hline
	NeuralCRF(C\&W)&0.9215&0.9306&0.9260&0.8956&0.8974&0.8965\\
    NeuralCRF(word2vec)&0.9132&0.9257&0.9194&0.8910&0.8896&0.8903\\
    NeuralCRF(Our model)&\textbf{0.9308}&\textbf{0.9451}&\textbf{0.9379}&\textbf{0.9047}&\textbf{0.9022}&\textbf{0.9034}\\
    CRF(character)&0.9099&0.9141&0.9120&0.8805&0.8769&0.8787\\
    CRF(character + radical)&0.9117&0.9153&0.9135&0.8816&0.8778&0.8797\\
	\hline	
\end{tabular}
\caption{Comparison of $F_1$ on the test set of CTB5 and CTB7.}\label{Table-compare}
\end{table*}
\paragraph{Results and Analysis}
\begin{figure}[htbp]
\begin{center}
\includegraphics[scale=0.95, width=0.95\columnwidth]{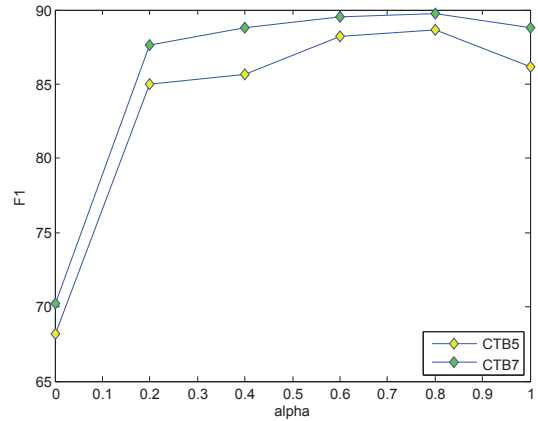}
\vspace{-1em}
\caption{$F_1$ on the development set of CTB5 and CTB7. $alpha$ is the weight of the context-based loss in our model.}
\label{Figure_fencialpha} %
\end{center}
\end{figure} %
Figure \ref{Figure_fencialpha} gives the experiment results of our radical-enhanced model on the development set.
The trends of our model are consistent on CTB5 and CTB7. Both performances increase sharply at $alpha=0.2$ because context information is crucial for this sequential labeling task yet not utilized in the purely radical-driven model~($alpha=0$). The best performances are achieved with $alpha$ in the range of [$0.6$,$0.8$]. Figure~\ref{Figure_alpha} and Figure~\ref{Figure_fencialpha} have different trends because of the different characteristics of the two tasks. For character similarity judgement, radical is dominant because it reflects the character category information. But for Chinese word segmentation, contexts also play an important role.

We compare our radical-enhanced model~($alpha$$=$$0.8$ tuned on the development set) with C\&W model and Word2Vec in the framework of Neural CRF. Table~\ref{Table-compare} shows that, our model obtains better $P$, $R$ and $F_1$ than C\&W and word2vec on both CTB5 and CTB7. One reason is that the radial-enhanced model is capable to capture the semantic connections between characters with the same radical, which usually have similar semantic meaning and grammatical usage yet not explicitly modeled in C\&W and word2vec. Another reason is that, the embeddings of lower-frequent characters are typically not well estimated by C\&W and word2vec due to the lack of syntactic contexts. In the radical-enhanced model, their radicals bring important semantic information thus we obtain better embedding results.
We also compare with two CRF-based baseline methods. $CRF(character)$ is the use of linear-chain CRF with character as its feature. In $CRF(character+radical)$, we utilize the radical information and the character as features with linear-chain CRF. Results of $CRF(character)$ and $CRF(character+radical)$ show that simply using radical as feature does not obtain significant improvement. Our radical-enhanced method outperforms two CRF-based baselines on both datasets, which further verifies the effectiveness of our method.
\section{Conclusion}
In this paper, we propose to leverage radical for learning the continuous representation of Chinese characters. To our best knowledge, this is the first work on utilizing the radical information of character for Chinese computational processing.
A dedicated neural architecture with a hybrid loss function is introduced to effectively integrate radical information for learning character embedding. Our radical-enhanced model is capable to capture the semantic connections between characters from both syntactic contexts and the radical information. The effectiveness of our method has been verified on Chinese character similarity judgement and Chinese word segmentation. Experiment results on both tasks show that, our method outperforms two widely-accepted embedding learning algorithms, which do not utilize the radical information.
\bibliographystyle{acl}
\bibliography{ref}

\end{document}